\pdfoutput=1

\documentclass[11pt]{article}

\usepackage{acl}

\usepackage{times}
\usepackage{latexsym}

\usepackage[T1]{fontenc}

\usepackage[utf8]{inputenc}

\usepackage{microtype}

\usepackage{graphicx}
\usepackage{booktabs}
\usepackage{pgfplots}
\usepackage{amsmath}
\usepackage{float}
\usepackage{multirow}
\title{SMPrompt: Semantic-driven Augmented Templates and Multiple Mappings for Effective Prompt Learning}

\author{Jinta Weng, Yue Hu,  Jing Qiu, Zhihong Tian, Heyan Huang \\
 University of Chinese Academy of Sciences, Institute of Information Engineering\\
 Beijing Institute of Technology \\
  Guangzhou University \\
  \texttt{wengjinta@iie.ac.cn} \\ \\}

\begin{document}
\maketitle
\begin{abstract}
 The effectiveness of prompt learning has been demonstrated in different pre-trained language models. By formulating suitable templates and choosing representative label mapping, it can be used as an effective linguistic tool. However, finding suitable prompt always need to multiple experimental attempts or appropriate vector initialization, while is more common in few-shot learning tasks. 
Therefore, we try to construct the prompt from semantic perspective and propose the SMPrompt -- \underline{S}emantic-driven augmented \underline{P}rompts with \underline{M}ultiple mappings. Specifically, two prompts generated from the semantic dependency tree (Dep-prompt) and task-specific metadata description (Meta-prompt) are firstly introduced. What is more, a multiple mappings strategy (Ml-prompt) is proposed to enhance the model's attention on different mapping tokens. Our results show that the proposed model achieves the state-of-the-art performance of prompt learning in few-shot text classification tasks\footnote{We will open source all codes for fast prompt design.}, which prove that the SMPrompt could assume as a faster knowledge proving tool of PLMs.


\end{abstract}

\section{Introduction}

Pre-trained language models (PLMs) have widely changed the research paradigm of natural language processing in recent years \cite{2018bert}. 
With fine-tuning parameters and more expressive linguistic features, PLMs are able to show better effectiveness in downstream NLP tasks. Ideally, linguistic knowledge and its semantic relations are re-parameterized and motivated by fine-tuning process. Thus, through the proper design of external prompt, the PLMs' knowledge can be greatly extracted. Motivated by this, different model designs and promoting methods, like GPT-3's demonstration learning \cite{Brown2020Language} and cloze question templates \cite{jiang2020can}, have show its powerful ability in few-shot learning tasks.
\par
\begin{figure}[htbp]
	\centering
	\includegraphics[width=0.5\textwidth]{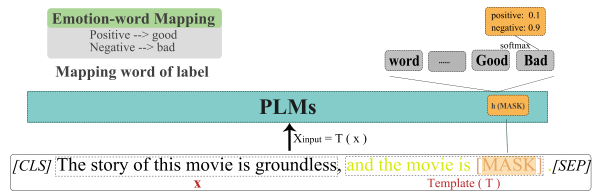}
	\caption{Using prompt learning method in emotion classification task. The input would add with a fixed template T (e.g., some words or sentence) and a $ [MASK] $ token to combine the real input, and the task prediction would then be mapped into the prediction of specif label-mapping word given the hidden value of mask token.}
	\label{fig:figure1}
\end{figure}

Comparing with different fine-tuning strategies, prompt-based fine-tuning use a more intuitive approach by simply defining a specif label mapping (also named as label verberlizer) and task-oriented template. As depicting in fig. \ref{fig:figure1}, a cloze question containing a masked token is added to original input x, and then the hidden representation of this ``[MASK]'' token is used to generate a token distribution on vocabulary degree. Consequently, the distribution of predefined label-mapping tokens are selected to realize the final prediction. 
Although prompt learning methods have achieved significant effect and development in few-shot learning tasks, related studies show minor change of the templates and label mapping could make the result widely different \cite{qiu_pre-trained_2020}. Thus, finding representative template and label mapping to construct an augmented prompt is still a worth-considering question. 
\par
Most researches try to explore suitable prompt from the aspect of its representation and model construction. For example, Schick and Schütze propose the PET and IPET models from the viewpoint of multi-templates learning and using expanded training dataset, while the hidden information of large unlabeled dataset are unavoidably reused \cite{schick2021exploiting}. In order to generate a suitable prompt, Chen et.al also design a violence searching method on selecting the suitable label mapping, which allows generating maximal hidden value of mask token on all inputs \cite{gao2020making}. However, Webson argue it more like a kind of relaxed RegEx rules fitting the task premise and hypothesis pairs of original corpora\cite{webson2021}. What is more, we also found that its searching space are explosive increase with label's number O($n^{n}$), and the generating templates and label mappings are based on existing experienced manual-defined template, which make prompt design still a specialist-first and computational-ability-first thing due to heuristic-discovery process. Different from the discrete prompt, recently, the continuous prompt learning methods have shown its effectiveness on P-tuning and prefix-tuning \cite{liu2021gptunderstand,li-liang-2021-prefix,shin2020autoprompt,schick2021exploiting}. By encoding the prompt in a continuous vector with limited length, PLMs are able to tuning the prompt vector and learn the hidden relations within template and original input. However, these methods makes the prompt-based learning becomes a block fine-tuning or adding-block fine-tuning method, thus lose its obvious feature on portability and interpretation while comparing with hard prompt. What is more, the suitable initialization of the continuous prompt is also essential to the continuous prompt on few-shot dataset. 
\par
In addition, Webson also found that the effectiveness of prompt learning depends mostly on the selection of label mapping \cite{webson2021}. However, most label mapping of the prompt-based method only use one token to represent a specific label. For example, only use word ``great'' to represent the emotion ``very happy'' and token ``good'' to ``happy'', or use ``yes'' to represent the positive sample and ``no'' for negative sample, which ignore the semantic similarity of mapping tokens, and restricts the multiple expression in the design of label mapping of real scenario. A nature question would be: Could we realize a more semantic or task-related prompt for constructing an interpretable prompt?


\par
In order to get close to this question, we proposed a \underline{T}ask-driven \underline{P}rompts with \underline{M}ulti-labels Mapping (TMPrompt). Our proposed paper aims to design more instructive prompts from the task-driven and sparse label-mapping viewpoint for augmented prompt tuning. In detail, the task-driven dependency-prompt (Dep-prompt) and the metadata-prompt (Meta-prompt) generating by semantic-dependency-tree and our task metadata description, are first developed to empower prompt's hidden knowledge. Subsequently, we propose a multi-labels mapping strategy (Ml-prompt) to allow the tuning process learning semantic diversity of different label mappings. 

The contributions of the proposed work in the domain of prompt-based few-shot classification are:
\begin{itemize}
	\item By defining an augmented prompt Dep-prompt constructing by suitable dependency-tree-based and POS-based filters, key information of original input is strengthened to empower the final prediction.
	\item Since PLMs have seen similar tasks and datasets in pre-training step, offering more types of metadata description could help the language models fine-tuning to more related space. Thus, Meta-prompt considering more task descriptions is developed to help the model finding the task-oriented knowledge on PLMs. 
	\item We offer a learning strategies on multi-labels mappings, which allows diversified expression on the label mapping, instead of an expert-designated or computation-prior label mapping. We change the goal of finding best label mapping into mutual decision based on controllable weight of multi-labels mapping. 
	
\end{itemize}
\par
Experiment shows the proposed TMPrompt achieves better result by using two semantic-driven prompts and applying more multi-labels mappings method. And the TMPrompt may give more semantic interpretation and technical guidance for fast prompt design in prompt-based fine-tuning model. In the following paper organization, we will briefly display the overview of current methods in prompt learning, and then introduce our proposed model and its effectiveness.




\section{Problem Definition}

Our main work is based on the design of textual prompt and prompt-based fine-tuning strategy with K-shot setting as \cite{gao2020making}. In this prompt-based fine-tuning method, the prompt generally consists of a manual template to generate a prompting input, a mask token for label prediction, and the label mapping that maps different labels into specified words like Fig. \ref{fig:figure1}. Subsequently, a per-trained language model is introduced to encodes the prompting input. In K-shot setting, the training corpus using for fine-tuning is construed by  k few samples of each label.
\subsection{Task Formulation}

We use D for the training set and $ D^{(i)}=(x^{i},y^{i})$ is the $i-th$ training pair of the current dataset. For text classification task, the task goal is predicting the label $y^{i}$ given $ x^{i} $.
After defining a transition template for current task, the original input $ x^{i} $ is transformed to prompting $ \overline{x}^{i} $ . If the template is `$ \textbf{x}, It\ is\ \textbf{[mask].} $', the transition is formulated as:
\begin{equation}
	\overline{x}^{i} = T(x^{i})= x^{i}. It\ is\ [mask].
\end{equation}
, where the \textit{[mask]} token would subsequently use to generate the word distribution over PLM's vocabulary.
\par
Some tokens in PLM's vocabulary $  V $ is chosen to represent each label and therefore construct a virtual label mapping. For example, in emotion classification task, we could use the PLMs token ``good'' stand for positive label, while use ``bad'' for negative label.
\begin{equation}	
	F(y): y_{t} \rightarrow v_{t}, y_{t} \in {Y}, v_{t} \in {V}
\end{equation}
, where $ t $ is the label index of label set $ Y $, and $ v $ represents the specif token of $ V $.  
\par
Given these transition, the target is predicting the label-mapping tokens $ v $ on \textit{[mask]} token.
\subsection{Model formulation}
In this section, we generate the hidden value $ h_{[mask]} $ on the  $ \overline{x}^{i}$ by a PLM M. Since each label has been represented by a specif token in label mapping, the hidden value $ h_{[mask]} $ would be used to create a token-level distribution over PLM's vocabulary by introducing a mapping matrix W:
\begin{equation}
	h_{v_{t}} = h_{[mask]}\cdot  W_{v_{t}} =  \cdot LM([mask]|	\overline{x}^{i}) \cdot W_{v_{t}} 
\end{equation}
, where the size of mapping matrix W is $R^{{|h}\times{len(V)|}}$, and $h_{v_{t}}$ is the value of PLM token distribution in token $v_{t}$.

Only the distributions of the mapping tokens $v_{t}$ would be retained and used to calculate the final prediction formulated in following equation:

\begin{equation}
	\begin{split}
		p(y^{t}|x^{i}) &= p(v^{t}| 	h_{[mask]}) \Rightarrow p(v^t |W \cdot h_{[mask]}) \\
		& 	\Rightarrow p(v^t | h_{v_{t}}) = \ln \dfrac{\exp(h_{v_{t}})}{\begin{matrix} \sum_{k=1}^{|Y|}\exp(h_{v_{k}}) \end{matrix}}
	\end{split}
\end{equation}
, where $|Y|$ is the label numbers, $v_{k}$ used to represent the token logit value of the k-th token in soft label mapping. 

The target of task classification thus transits into the prediction of tokens in label mappings and the final loss is formulated as:
\begin{equation}
	L=-\dfrac{1}{N}\sum_i\sum_{t=1}^{|Y|} y_{it}\log{p(y^{it}|x^{it})}
\end{equation}
, where $i$ is the index of the training pair $(x_{i},y_{i})$.

Since the label mapping function $ F(y) $ and template transition $  T(x) $ largely determines the performance of the final prediction, selecting suitable template $  T$ and correct label mapping method  $F(y)$ is essential for prompt-based fine-tuning method. 

\begin{table}[H]
	\centering
	\scalebox{0.9}{
		\begin{tabular}{|c|l|l|}
			\hline
			\textbf{Template} \  &\textbf{ Mapping} & \textbf{Acc.} \\
			\hline
			\multirow{2}{*}{It is [*mask*] } & {terrible,great}   & 92.1 \\ \cline{3-2}
			& {terrible,good}  & 90.9 \\ \hline
			\multirow{2}{*}{{This movie is [*mask*] }} & {terrible,great}  & 92.9 \\ \cline{3-2} 
			& {great,terrible} & 86.7 \\ \hline
	\end{tabular}}
	\caption{Different combination of template and label mappings in SST-2 emotion classification task. The i-th word of label mappings represent i-th label.}
	\label{tab:tabel0}
\end{table}
As shown in  table \ref{tab:tabel0} of promt-based fine-tuning in SST-2 tasks, diverse combinations of prompt and template design usually leads the result quite different. Therefore, finding the optimal template and mapping has become the priority for experts and computation.
\section{Methodology}
Therefore, the proposed model TMPrompt tries to exploit this question from task-driven template design and multi-labels mapping strategies. \par
\begin{figure*}[htbp]
	\centering
	\includegraphics[width=1\textwidth]{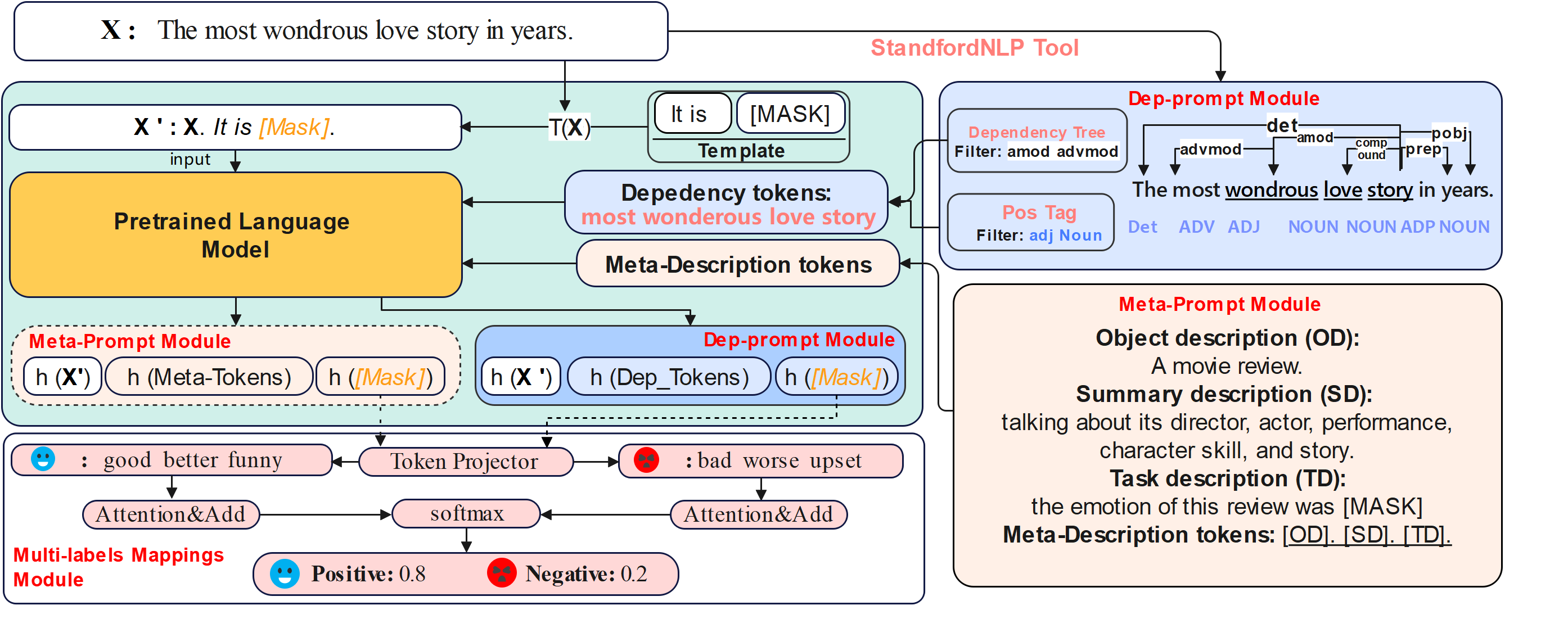}
	\caption{The proposed TMPrompt model. We task the emotion classification SST-2 task as example to explain the model. We use different colors to express the blocks of Dep-prompt, Meta-prompt, and ML-prompt.}
	\label{fig:model}
\end{figure*}\par

\subsection{\textbf{Dep-prompt: Dependency-based generative prompt}}
In terms of template design, we use more task-driven strategies to generate two different kinds of augmented prompts, the Dep-prompt considering the semantic-dependency-tree filters and the Meta-prompt from metadata description of current task. \par

\par
\begin{table}[H]
	\centering
	\begin{tabular}{p{2cm}p{5cm}}
		\toprule
		\textbf{Filter Type} &  \textbf{Filters Name} \\                   
		\midrule
		Word Class (POS)  &  amod $|$ advmod $|$ obj $|$  NN $|$ VBD $|$ VBZ $|$ VB  \\\midrule
		Dependency Tree  & amod (Adj) $|$ advmod (Adv) $|$ ROOT $|$ obj $|$ nsubj\\
		\bottomrule
	\end{tabular}
	\caption{Dep-filters. For more POS and dependency relation, please referring stanford dependency tree.}
	\label{tab:tabel2}
	
\end{table}
%

This kind of template design is based on the idea of reinforcement of task semantic, since some tasks are more related with specific dependency or word class. For example, adjectives or adverbs may contain more sentiment expression, and therefore if we could add the adjectives ``beautiful'' before the \textit{mask} token, the prediction of `mask' token would prefer the mapping token of `positive' emotion. In this way, text entailment task could also simplify into the special entity and relation comparisons in the context, which these entities and relations could be sieved by special Pos and dependency. 

Comparing with tradition GPT-3's in-context learning, these semantic tokens (eg. adjectives, nouns, entity and relation) contain more task-related features and less information interference in fine-tuning process.
Therefore, we use these semantic tokens to construct first type of augmented prompt named as Dep-prompt. 

We generate these semantic tokens sieving by the Pos filter and dependency tree filter by StandfordNLP Tagger \footnote[1]{https://nlp.stanford.edu/software/tagger.shtml}. 

\par
Only the words with specific POS type and dependency type are able to be chosen. Considering the tasks' distinction, different task may choose few specif dep-filters.  As Fig. \ref{fig:model} depicted, after defining Pos filters and dependency-tree filters, some specific tokens are selected to formulate a Dep-prompt and added into the original input. The template transition of Dep-prompt would be formulated as:
\begin{equation}
	\overline{x}^{i} = T_{dep}(x^{i})= x^{i}.\ \underline{x^{dep}}.\ It\ is\ [mask]
\end{equation}
\par
For example, we use the words belong to ``amod'' dependency filters or ``NN'' Pos filter to construct a dep-prompt for emotion classification task, since the result of emotion classification is more related about the ``amod'' relation and noun-class word. Note that the length of $ x^{dep} $ is up to the filter number, while using a Pos filter may add one token and two for selecting one dependency tree filter. 

However, randomly using all tokens also result in these problems: (a) The length of dependency tree may exceed original input of PLM, and thus introduces more useless and interference token information ; (b) Different tasks seem to be sensitive about different tokens with specif dependency type and POS tag. 
Therefore, as Tab. \ref{tab:tabel2} depicted, we first construct two different types of filters on dependency and Pos type to restrict the available dependency types and POS tag for constructing the augmented prompts.

What is more, we introduce a grid searching strategy to automatically finding the most suitable filter in filter candidates. We have list some dep-filters generated by auto-searching method in Appendix \ref{sec:appendixB} to realize task adaption and quick setting. For unknown tasks or datasets, user could also use our auto-searching tools to find the suitable filter.

\par

\subsection{Meta-prompt: Metadata Description Prompt.}
{\tiny }
The effectiveness of prompting learning method have shown PLM is a parameterized knowledge base of training corpus. Therefor, prompt-based fine-tuning process not only optimizing huge parameters to adapt in downstream task, but also finding a best parameterized path to access similar dataset. Motivated by this, we introduce another augmented prompt Meta-prompt to construct the template from the metadata description of current task, such as label information, task goal, and other metadata information.

As following equation \ref{eq:meta}, our defined Meta-Prompt is composed of object description (OD), summary description (SD), and task description (TD) with a mask token. 
\begin{equation}
	\overline{x}^{i} = T_{dep}(x^{i})= x^{i}.[Od][Sd][Td][MASK]
	\label{eq:meta}
\end{equation}
\par
We list a example of meta-prompt of SST-2 emotional classification tasks in Tab. \ref{tab:tabel2}. For more meta-prompt of different tasks, please referring in appendix \ref{sec:appendixC}.
\begin{table}[H]
	\scalebox{0.98}{
	\centering
	\begin{tabular}{p{7cm}}
		\toprule
		\textbf{Task name (labels)}\\  -Metadata description template \\                   
		\midrule
		\textbf{Example: SST-2 Task} (positive / negative) \\ \textbf{- OD}: A movie review \\ \textbf{- SD}: Talking about its director, actor, performance, character skill, and story. \\ \textbf{- TD}: The emotion of this review was [MASK]  \\
		\midrule
	\end{tabular}
}
	\caption{Our Meta-prompt combining by object description (OD), summary description (SD), and task description (TD) with a `mask' token.}
	\label{tab:tabel3}	
\end{table}


Since PLMs may see similar tasks and corpus in the pre-training process, we could use more metadata description about current task to help the model recalling its knowledge memory quickly, instead of merely use a cloze question or demonstration example like GPT-3. What is more, the metadata description could help the PLMs fine-tuning in more narrow and similar semantic space, for the reason that the metadata description template introduces more task-related information.
\subsection{ ML-Prompt: Multi-labels Mapping}
In terms of current label mapping, only one PLM's token could be seemed as the representative soft label. We argue that this restriction limits the model's diversity and it is hard to find a representative label mapping while considering difficult tasks and calculation. What is more, its not easy to find a suitable label mapping by exhaustive searching, since the label mapping choice and search space would grow exponentially with the number of labels and vocabulary. Also, the limited number of label mapping loss its robustness and transferability in similar task. In order to enable the model learn more distinct information from multiple tokens and reduce the difficulty of the definition of label mapping, a multi-labels mapping method is introduced, and we named it Ml-Prompt.

\par
In detail, each of labels would map to more than one token in vocabulary:

\begin{small}
\begin{equation}	
	F(y): y_{t} \rightarrow \{v^{1}_{t},v^{2}_{t},...,v^{n}_{t}\}, y_{t} \in {Y}, v^{n}_{t} \in {V}
\end{equation}
\end{small}
, where n is the number of mapping tokens of specif label t. For example, in SST-5 emotional classification task, traditional label mapping may use word set $\{disgusting, awful, horrible, ridiculous, boring\}$ as the identified tokens for the label `\textit{very negative}'. Our detail multi-labels mapping of different tasks is listed in appendix \ref{sec:appendixD}.
\par
Considering different label mapping tokens may introduce differential gain, the multi-labels mapping matrix M for learning different weights of multiple tokens is defined. M is a weighted matrix that initialize by Xavier normally distribution.


Thus, the prediction considering weighted tokens is formulated as: 
\begin{equation}
	\begin{split}
		p(y^{t}|x^{i}) &\Rightarrow p(v^t | h(x^{i})) \\
		&= \dfrac{ exp(M_{t}  \cdot h(x^{i})+ b_{t}) }{\begin{matrix} \sum_{t^{'}}  exp( M_{t^{'}}  \cdot h(x^{i})+ b_{t^{'}})\end{matrix}}
	\end{split}
\end{equation}
, where the size of mapping matrix M is $R^{{|n}\times{t|}}$.
What is more, in order to further improve the robustness of different numbers of label mappings, we also consider the attention on different numbers of mapping tokens in pre-defined label mappings. The final prediction of each label $y^{t}$ would be: 
\begin{small}
\begin{equation}
	\begin{split}
		p(y^{t}|x^{i}) &\Rightarrow \dfrac{1}{B}\sum_{B} p(v^t | h_{v_{t}}) \\
		&= \dfrac{1}{B}\sum_{B} \dfrac{exp(M_{t}^{B}  \cdot h(x^{i})+ b_{t}^{B}) }{\begin{matrix}\sum_{t^{'}} exp(M_{t^{'}}^{B}  \cdot h(x^{i})+ b_{t^{'}}^{B})\end{matrix}} 
	\end{split}
\end{equation}
\end{small}
, where B is the number of using label mappings. 

If we set B=4, it represents prompt tuning with four different multi-labels mappings. The final loss thus formulated as:
\begin{equation}
	L=-\dfrac{1}{N}\sum_{i}^{N}\dfrac{1}{B}\sum_{B} y_{i}\log{p(v^{t}|x^{i})}
\end{equation}

From the depicting formula, the task prediction thus could consider different attentions on the level of multiple tokens of each label and the level on different types of label mappings. The Ml-Prompt not only allow the model learning from more relaxation tags instead of a specif label, but also could learn minor distinction of different multi-labels mappings. 
\begin{table*}[htbp]
	\scalebox{0.9}{
	\centering
	\begin{tabular}{lccccc}
		\toprule
		\textbf{Baselines} \  &\textbf{ TREC (acc)} & \textbf{SNLI (acc)} &\textbf{ QNLI (acc)} & \textbf{SST-5 (acc)}& \textbf{ SST-2 (acc) } \\
		\midrule
		Majority         $ \star $                   & 18.8      & 33.8      & 49.5 & 50.9      & 23.1      \\
		prompt-based zero-shot learning$ \star $     & 32.0      & 49.5      & 50.8        & 35.0   & 83.6    \\
		GPT3-in-context-learning$ \star $            & 26.2(2.4) & 47.1(0.6) & 53.8(0.4)    & 30.6(0.9) & 84.8    \\
		fine-tuning$ \star $                         & 26.2(2.4) & 48.4(4.8) & 60.2(6.5)  & 43.9(2.0) & 81.4(3.8) \\
		\midrule
		LMBFF$ \star $ & 84.8(5.1) & 77.1(3.9) & 64.5(4.2) & 46.1(1.3)  & 92.1(1.1) \\
		P-tuning$ \star $ & 84.8(5.1) & 77.1(3.9) & 64.5(4.2) & 46.1(1.3)  & 92.1(1.1) \\
		Prefix-tuning$ \star $ & 84.8(5.1) & 77.1(3.9) & 64.5(4.2) & 46.1(1.3)  & 92.1(1.1) \\
		
		\midrule
	\textbf{	Our Methods}      &   &  &    &     &  \\
	
	-	Dep-prompt $\dagger$   &\textbf{ 87.2 (3.4)} & \textbf{77.6 (1.7)} &\textbf{ 65.9 (3.3})  &  \textbf{48.1(1.0)} & 91.8   \textbf{(0.7)} \\
		- Meta-prompt $\dagger$  & \textbf{85.0(2.0)} & \textbf{77.2 (1.4)} & 61.0 (5.8)    &  \textbf{50.4(1.3)} & 91.9  \textbf{ (1.5)}\\
			- ML-prompt $\dagger$      & 83.4 (3.4) & 72.5 (3.3) &\textbf{68.5 (3.2)}    &    46.0 (1.7)   & 89.3  (4.2)  \\
		\midrule
		\textbf{TMprompt}  & 84.8(5.1) & 77.1(3.9) & 64.5(4.2) & 46.1(1.3)  & 92.1(1.1) \\
		\bottomrule

	\end{tabular}
}
	\caption{Main result of the TMPrompt. The results of all experiments are evaluated by selecting the mean and variance of accuracy on five different segmented training datasets and same testing dataset. The result of existing baselines ($ \star $) are using from the \cite{gao2020making}, and our proposed baselines is Dep-prompt($\dagger$) , Meta-prompt($\dagger$) , and ML-prompt($\dagger$).}
	\label{tab:tabel4}
\end{table*}

%
%

\section{Experiments}
In this section, we present the effectiveness of proposed method during the sentiment classification, question classification, and semantic entailment tasks based on few-shot setting.
\subsection{Datasets}
We choose sentiment classificaition, question classification and natural languge interfence dataset to present our proposed approach. In sentiment classifcation, we choose two representative SST Task, sst-2 and sst-5. The goal of SST dataset is given a movie review to predict its emotion. The label of sst-2 task is constitute of postive and negative tags, while in sst-5 task are tags of `very postive', `positive', `netural', `negative', and `very negative'. In question classification dataset, TREC-6 task is chosen to explore the task's understanding on QA. TREC task aim to predict on the six question types given an English question text. In nature language inference task, we use SNLI and QNLI dataset to test the model's ability on judging whether the previous sentence could imply the semantics of the next sentence in the text. For more information about datasets we listed in appendix \ref{sec:appendixA}.
\subsection{Experiment Setting}
We use RoBERTa-large as our pre-training language model. Our experiments are developed in NVIDIA V100 32GB (also could run in 1080ti with low batch size). In the training process, we develop many experiments on different batch sizes \textit{bs=4,8,16}, learning rate \textit{lr={1e-5,2e-5,5e-5}}. For the reason that some study have shown the minor difference on few-shot dataset could lead differentiated results, we use five different sub-sets with the same sampling size. Also, in order to control the fairness of experiments, we use the mean score and variance of the prediction result on different subsets instead of the highest score.
To satisfy the few-shot learning in PLMs, following the setting in Gao's work, we pick five different K-shot sub-datasets and each sub-dataset is constructed by K=16 training pairs on each type of labels \cite{gao2020making}. For example, the SST-2 emotion classification task with two classes needs to construct five different training sets with the size of 32 and validation set with the same size of 32, while the testing size would use original size of the SST-2 task, without any other data setup. These sub-datasets are training and predicting singly. Their training process and prediction result would be recorded by different batch sizes and learning rates.
All training pair in same task would use the same template transition in proposed prompting strategies to construct the real input.
Finally, we choose the best result of each sub-dataset on different hyperparameters, and integrate the best result on the aspect of sub-dataset.

We chose the evaluation criteria in integrating the prediction of sub-datasets as the average accuracy and variance of different sub-dataset, since it reveal the overall performance and variation of current task. 

\par
\subsection{Baselines}
We compare with a number of baselines in related works and set our baselines with Majority method (merely select the majority class as prediction), fine-tuning method \cite{Liu2019RoBERTaAR}, prompt-based zero-shot learning, GPT3-in-context-learning \cite{Brown2020Language}, and LMBFF model (The method detail refers to Gao's work  \cite{gao2020making} ). Also, the proposed three methods named as Dep-prompt, Meta-prompt, and Ml-prompt of the TMPrompt are also compared.
%

\begin{figure*}[htb]
	\pgfplotsset{width=17cm, height=5cm, compat=1.5}
	\scalebox{0.95}{
	\begin{tikzpicture}
		\begin{axis}
			[
			ybar, 
			enlargelimits=0.1,
			legend style={at={(-0.11,1.0)}, 
				anchor=west,legend columns=1},   
			ylabel={\#Accuracy}, 
			symbolic x coords={LMBFF, JJ, WP, WRB, NNP, WDT, NN, amod, advmod, ROOT, obj, nsubj},
			xtick=data,
			nodes near coords,
			nodes near coords align={vertical},
			]
			\addplot coordinates {(LMBFF, 64.5) (JJ, 68.1)  (WP, 68.7 )(WRB, 68.1 ) (NNP, 68.1 ) (WDT, 67.7 ) (NN, 66.5) (amod, 68.1) (advmod, 67.1) (ROOT, 70.8) (obj, 66.2) (nsubj, 65.6)};
			\addplot coordinates {(LMBFF, 77.1) (JJ,74.9)  (WP, 75.7 )(WRB, 76.2 ) (NNP, 77.6 ) (WDT, 77.4 ) (NN, 74.4) (amod, 73.5)  (advmod, 76.2 ) (ROOT, 75.5 ) (obj, 68.2 ) (nsubj, 76.1) };
			
			\legend{SNLI, QNLI}
		\end{axis}
		
	\end{tikzpicture}
}
	\caption{The effectiveness of different using dependency filters in Dep-prompt. The first column marked by \textbf{LMBFF} is the effectiveness of prompt-based fine-tuning baseline of \cite{gao2020making}.}
	\label{fig:figure3}
\end{figure*}
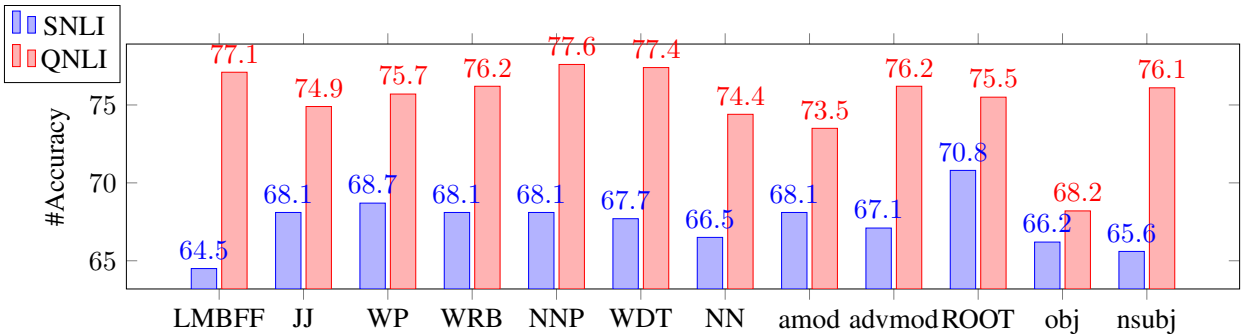

%
\subsection{Main Result}
The main result of different proposed methods are depicted in Tab. \ref{tab:tabel3}. As the result shown, the proposed TMPrompt show better performance on different types of task in following aspects: (a) Comparing with the fine-tuning and GPT-3 in-context learning method, our TMPrompt could get better performance on different tasks, for the reason that our TMPrompt adopt more instructive prompt-based fine-tuning method like LM-BFF; (b) Owing to more keyword information generating by semantic dependency tree, the Dep-prompt template design could improve the performance of the TREC, QNLI and SST-5 task with higher accuracy and lower variance; (c) The performance of TREC, SNLI, and SST-5 tasks indicate that suitable Meta-prompt design of current task could help the model get more information on task level. (d) The result of Ml-prompt design point out that multiple-labels mapping also could obtain better effect on QNLI task since it could consider more distinction and similarity from different multi-labels mappings, which allow controllable label relaxation and reduce the searching difficulty of specifying label mappings.
\par
Also, the effectiveness of ML-prompt reveals that, comparing with one mapping token for each label, using more tokens may bring little interference although it gives the prompt designer more choices on defining label mappings. What is more, although we use more semantic-driven template and multi-labels mappings to construct augmented prompt, all the proposed design seem insensitive to binary emotional classification task SST-2. 

\subsection{Choice of Dep Filters}
Owing to using different dep filters may add new information to original input, we select wo text entailment tasks, QNLI and SNLI task to explore the effectiveness of identical filter names. As Fig. \ref{fig:figure3} depicted, comparing with the LMBFF model, the result shows that the POS-type filter with ``NNP'' and ``'WDT' could  reach  better results. In QNLI task, all different types of proposed filters are able to reach the state-of-the-art result, which reveals that the dep-prompt could empower the performance of text inference task in some degree.
\subsection{Choice of Meta-prompt}
\begin{table}[htb]
	\centering
	\scalebox{0.95}{
	\begin{tabular}{lccccc}
		
		\toprule
		\textbf{Combinations}   &\textbf{ Acc.} & \textbf{Variance} & \textbf{Median} \\
		\midrule
		OD+SD+TD w/0  & 46.1    & 1.3 & 46.4  \\\midrule
			TD  & 47.5   & 2.3 & 48.5  \\
		OD  & 48.2  & 2.5 & 48.4  \\
		SD         &48.3    & 2.1 & 47.1  \\\midrule
		SD+TD	  & 48.7$ \uparrow $  &  0.6$ \downarrow $   & 48.6$ \uparrow $  \\
		OD+TD        & 48.8$  \uparrow $     & 1.0$ \downarrow $ & 48.6$ \uparrow $  \\
		OD+SD       & 50.1  $ \uparrow $      & 1.4$ \downarrow $ &  49.8$ \uparrow $ \\\midrule	
		OD+SD+TD         & 47.3 $ \downarrow $   &  5.4  $ \uparrow $ & 50.0$  \uparrow $ \\\midrule
		
		\textbf{Esemnble}   & 49.5 &1.7  &50.3$  \uparrow $\\
		\bottomrule
		
	\end{tabular}
}
	\caption{The effectiveness of different types and combinations of description of Meta-prompt in SST-5 emotion classfication. The ensemble result is generated from the joint inference of all combinations of Meta-prompt.}
	\label{tab:tabel5}
\end{table}
In order to explore different types of task description in Meta-prompt, we further develop ablation experiments on different types of meta-prompt in SST-5 emotion classification task. In detail, we use different experimental combinations constructed by three types of meta-prompt, object description (OD), summary description (SD), and task description (TD). Implementation details and using template are in Appendix \ref{sec:appendixC}. As the results shown, all of these three types of description are able to motivate the PLMs to generate more suitable predictions. With the increase in the type of description, the accuracy and median value of current task could be improved and its variance are decreased. It indicates that using these three types of description and their combination to construct a prompt are able to help the PLMs find the task-related knowledge embedded in PLMs and increase the effectiveness of current task. However, though the median value is incresing, the accuracy begins to reduce and its variance is increased when we try to combine these three types of description. Therefore, using more types of descriptions sometimes brings more interference when the length of augmented tokens are much longer than the original sentence. Also, it reveals that we should keep a balance between the length of these task descriptions and original input, while some methods used for representing these task description are also should be introduced.

\subsection{Multi-labels Mapping}

\begin{table}
	\centering
	\scalebox{0.95}{
	\begin{tabular}{p{1.8cm}p{2.3cm}p{2.3cm}}
		\toprule
		\textbf{\textbf{Mapping Numbers} }   & \textbf{QNLI} & \textbf{SST-5} \\
		\midrule
		1 &  62.5 (3.2)  & 44.1 (1.3)   \\\midrule
		2 & 62.6 (6.7)  & \textbf{44.7 (4.8) }  \\\midrule
		3 & \textbf{66.3 (5.2)}    & 43.1 (4.6)  \\\midrule
		4 & 65.5 (5.1)   &  44.0 (2.7)  \\\midrule
		5 & 	 66.1 (5.3)              &43.7 (1.7)
		\\\midrule
		\textbf{Esemnble}& 68.5 (3.2) & 46.0 (1.7)\\
		\bottomrule
	\end{tabular}
}
	\caption{The comparison of different label mappings with increasing numbers of tokens in multi-labels mapping. We still use the average accuracy and variance of different mapping and the `ensemble' result is generated from the joint inference by the combination of all multi-labels mappings.}
	\label{tab:tabel6}
\end{table}
We further explored the effectiveness of Ml-prompt by controlling the number of tokens corresponds to a label, and the results of QNLI and sst-5 task are depicted in table \ref{tab:tabel6}. As the number of mapping tokens increasing, the result shown that our proposed Ml-prompt could gain higher accuracy and lower variance on QNLI task. However, the result also shown that the suitable number of mapping tokens is 2-4. What is more, we have gathered the prediction of many multi-labels mappings ensemble with the difference on increasing mapping tokens \cite{schick2021exploiting}. The result shown that the effect of Ml-prompt would be further improved by integrating the result of all multi-label mappings.

\section{Conclusion}
Prompt-tuning methods have shown its effectiveness on different few-shot tasks. However, there are still exist expert-designated and computation-prior problems. In order to construct more instructive prompts and decrease the effort on finding the suitable label mapping, we proposed the prompt learning model--TMPrompt. Considering task-concerning semantic dependencies and task-driven description, we firstly introduce two prompt augmenting strategies, the Dep-prompt generating by two kinds of task-specific filters and the Meta-prompt construing by metadata description of current task. What is more, a multi-labels mapping method helps prompt-based fine-tuning gain from different mapping tokens is proposed. Our results show that TMPrompt could require better improvement on five few-shot classification task including emotion classification, question classification, and nature language inference tasks, which reveals the proposed two augmented prompts and the multi-labels mapping are able to assume as a better knowledge probe of PLMs. 

However, the TMPrompt model still exists some limitations: (a) The proposed prompt designs are based on experiential test on different dependent filters and task-oriented metadata description; (b) Though prompt-based fine-tuning method could achieve fast fine-tuning, an automotive and fast searching these tokens should be considered. Our further work would concentrate more on this and offer more technical guidance of TMPrompt.
\newpage
\bibliography{anthology,custom,ijcai22}
\newpage
\appendix
\setcounter{table}{0}
\renewcommand{\thetable}{A.\arabic{table}}
\label{sec:appendix}
\section{Data}
\label{sec:appendixA}
For SST5 \cite{socher2013recursive} and TREC \cite{voorhees2000building}, we use their official test sets. For SNLI \cite{bowman2015large} and datasets
from GLUE \cite{wang2018glue} , including SST2 \cite{socher2013recursive}, MNLI (Williams et al., 2018), QNLI \cite{rajpurkar2016squad}, we follow \citet{zhang2020revisiting} and use their original development sets for testing.  All of our utilizing dataset statistics are depicted in Tab. \hyperref[Tab:A1]{A.1}.
 \begin{table*}[htb]
 	\scalebox{0.75}{
 	\begin{tabular}{llllllll}
 		\hline
 	\textbf{	Category   }                      &\textbf{ Task}  & \textbf{|Y|} & L     & \#Train & \#Test & Type                         & \textbf{Labels}(Classification tasks)                  \\ \hline
 		\multirow{3}{*}{Single Sentence} & SST-2 & 2   & 19    & 6,920   & 872    & Sentiment                    & positive, negative                            \\
 		& SST-5 & 5   & 18    & 8,544   & 2,210  & sentiment                    & v. pos., positive, neutral, negative, v. neg. \\
 		& TREC  & 6   & 10    & 5,452   & 500    & Question cls.                & abbr., entity, description, human, loc., num  \\ \hline
 		\multirow{2}{*}{Sentence Pair}   & QNLI  & 3   & 11-30 & 104,743 & 5,463  & Nature Language Inteference  & entailment, not entailmen                     \\
 		& SNLI  & 2   & 14-8  & 549,367 & 9,842  & Nature Language Interference & entailment, neutral, contradiction            \\ \hline
 	\end{tabular}
}
\caption{The datasets evaluated in our work. |Y|: classes Number for classification tasks. L: average words in input sentence(s). In our few-shot experiments,  we also sample $ D_{train} $ and $D_{dev}$ of K × |Y| examples from the original training set. We use the dev set with the same number of samples in train set, while the test sets are chosen from the official datasets without any change.}
\label{Tab:A1}
 \end{table*}

\section{Choose Fine-grained Filters Name for Dep-prompt}
\label{sec:appendixB}
\setcounter{table}{0}
\renewcommand{\thetable}{B.\arabic{table}}

In order to find the suitable task-related filters for constructing a augmented prompt for different task, we conducted a pre-experiment on two different kinds of filters under different tasks. In Pos Filter, we use the word class generating from \href{https://nlp.stanford.edu/software/tagger.shtml}{POS Tagger} $\footnote[1]{https://nlp.stanford.edu/software/tagger.shtml}$ and select the `NN', `WDT', `NNP', `WRB', `WP', `NNP' and `JJ' as the POS filters. In Dependency Filter, we choose dependency relation `nsubj', `amod', `advmod', `ROOT', and `obj' named by  \href{https://downloads.cs.stanford.edu/nlp/software/dependencies_manual.pdf}{Standford Dependencies Representation}. 

\subsection{Setting}
Since our proposed method is based on the prompt-based fine-tuning, the label mapping and prompt template are predefined. We are based on the principle of minimum improvement and follow the formal \citet{gao2020making} setting. Our final label mappings and templates used in Dep-prompt section are depicted in Tab. \hyperref[Tab:B1]{B.1}.

We use RoBERTa-large as our pre-training language model in this section. Our experiment is developed in 1080ti with low batch size). In the training process, we set the batch size to \textit{bs=4} and learning rate \textit{lr={1e-5}}. We still use the average accuracy and variance value to evaluate the performance. 
\begin{table*}[htb]
	\scalebox{0.75}{
	\begin{tabular}{l|ll}
		\hline
		\textbf{Task} & \textbf{Label Mapping}                                                          & \textbf{Template}                             \\ \hline
		SST-2         & \{'0':'good','1':'bad'\}                                                        & *cls**sent\_0*\_It\_was*mask*.*sep+*          \\
		SST-5         & \{`contradiction':`No',`entailment':`Yes',`neutral':`Maybe'\}                   & *cls**sent\_0*\_This\_movie\_was*mask*.*sep+* \\
		TREC          & \{0:`Description',1:`Entity',2:`Expression',3:`Human',4:`Location',5:`Number'\} & *cls**mask*:*+sent\_0**sep+*                  \\
		QNLI          & \{`not\_entailment':`No',`entailment':`Yes'\}                                   & *cls**sent-\_0*?*mask*,*+sentl\_1**sep+*      \\
		SNLI          & \{`contradiction':`No',`entailment':`Yes',`neutral':`Maybe'\}                   & *cls**sent-\_0*?*mask*,*+sentl\_1**sep+*      \\ \hline
	\end{tabular}
}
\caption{The label mapping and template in the Dep-prompt. |Y|: classes Number for classification tasks. L: average words in input sentence(s). In our few-shot experiments,  we also sample $ D_{train} $ and $D_{dev}$ of K × |Y| examples from the original training set.}
\label{Tab:B1}
\end{table*}

\subsection{Effectiveness}
As shown in Tab. \hyperref[Tab:B2]{B.2} and Tab.\hyperref[Tab:B3]{B.3}, different type of filters used for constructing the augmented prompt Dep-prompt are able to enhance the model's effectiveness. In POS filter selection, adding the tokens with specific word class to construct an augmented prompt is effective. 'WP'  is beneficial to SST-2 task and QNLI task, while the tokens with POS `NNP' are effective in TREC and SNLI task. Also, the tokens with word class 'WRB' are able to enhance the SST-5 performance. In dependency tree filters, the `nsubj' relation is useful for TREC task, and the `ROOT' dependency is effective for QNLI, SNLI, and SST-5 tasks. What is more, it seems the tokens in our selecting dependency relation is not obvious to SST-2 task.

\begin{table*}[htb]
	\begin{tabular}{lllllllll}
		\hline
		\multirow{2}{*}{\textbf{Task}} & \multicolumn{8}{c}{\textbf{POS Filter}}                                                  \\ \cline{2-9} 
		& Orgin & NN         & WDT        & NNP        & WRB        & WP         & NNP        & JJ         \\ \hline
		SST-2                   &     89.3(1.3)   & 81.6 (6.4) & 84.1 (4.7) & 86.9 (2.6) & 85.6 (4.7)& \textbf{ 88.1 (1.8)} & 84.1 (4.7) & 83.5 (5.6) \\
		TREC                    &     82.8(3.7)  & 83.2 (3.7) & 84.4 (4.9) & \textbf{87.2 (3.4) }& 84.0 (5.2) & 84.0 (5.2) & \textbf{87.2 (3.4)} & 79.4 (6.1) \\
		QNLI                     &     64.5(3.2)  & 66.5 (3.6) & 67.7 (2.5) & 68.0 (2.5) & 68.0 (1.8) &\textbf{ 68.7 (2.1)} & 68.0 (2.5) & 68.0 (1.8) \\
		SNLI                      &   74.1(3.9)   & 74.4 (4.1) & 77.4 (2.0) & \textbf{77.6(1.9)}  & 76.2(2.5)  & 75.7(2.1)  & \textbf{77.6(1.9)}  & 74.9(2.4)  \\
		SST-5                      &   43.1(1.7)   & 49.4 (0.8) & 47.8 (3.1) & 48.4 (1.6) & \textbf{50.4 (1.3)} & 49.2 (1.3) & 49.2 (2.7) & 49.2 (2.7) \\ \hline
	\end{tabular}
\caption{The comparison of different POS filter. We use the original manual template adding by tokens of specific POS filter and label mapping defined by in Tab. \hyperref[Tab:B1]{B.1.}. The `Orgin' mark the experiment is developed based on prompt-based fine-tuning without any change.}
\label{Tab:B2}
\end{table*}

%
\begin{table*}[htb]
	\centering
	\begin{tabular}{lllllll}
		\hline
		\multirow{2}{*}{\textbf{Task}} & \multicolumn{6}{c}{\textbf{Dependency Filter}}                                      \\ \cline{2-7} 	    &   Orgin & nsubj      & amod       & advmod     & ROOT       & obj                \\ \hline
		SST-2                     &  89.3(1.3)        & 84.4(9.1)  & 82.1(10.5) & 85.4(7.3)  & 86.6(6.6)  & \textbf{87.9(4.3)} \\
	TREC                            & 82.8(3.7)   &\textbf{ 85.5 (5.5)} & 81.9 (5.1) & 79.9 (7.1) & 81.3 (7.5) & 84.4 (6.6)         \\ 
		QNLI                       &    64.5(3.2)     & 61.9 (3.9) & 60.6 (2.6) & 60.0 (3.6) & \textbf{64.1 (3.5)} & 62.7 (3.3)         \\
		SNLI                         &   74.1(3.9)    & 69.6 (6.3) & 67.6 (4.7) & 68.6 (6.5) &\textbf{ 70.8 (5.3)} & 65.2 (8.4)         \\	
			SST-5                       &   43.1(1.7)   & 40.9 (5.0) & 41.6 (2.4) & 40.9 (2.7) & \textbf{42.4 (3.2)} & 39.2 (6.1)         \\
			\hline
	\end{tabular}
\caption{The comparison of different Dependency Tree filters. We use the original manual template adding by tokens of specific dependency relation and label mapping defined by in Tab. \hyperref[Tab:B1]{B.1}}
\label{Tab:B3}
\end{table*}

\subsection{Different K-shot Setting}
Since related researches show that the K-shot setting is the key point for few-shot learning experiment, we also conduct a comparative experiment to explore the proposed Dep-prompt on different K. In detail, we construct different datasets which K=8,16,32,64,128 and 160 on TREC and SST-5 tasks. We also fix batch size bs=4 and learning rate r=1e-5, and max step=1000. The result is shown in  Tab. \hyperref[Tab:B4]{B.4}, which reveals the K=16 are benefit more than other K-shot setting.
\begin{table*}[htb]
	\scalebox{0.8}{
	\begin{tabular}{lllll}
		\hline
		\multirow{2}{*}{K-shot } & \multicolumn{4}{c}{Task (Filter Type)}                                                                                             \\ \cline{2-5} 
		& \textbf{TREC (Dependency Filter)} & \textbf{TREC (POS filter)} & \textbf{SST-5 (Dependency Filter)} & \textbf{SST-5 (POS filter)} \\ \hline
		8                       & 74.2 (+0.0)          & 78. 3(+0.0)             & 46.7 (+0.0)                         & 47.0 (+0.0)     \\
		16                      &\textbf{ 83.8 (+9.6)  }      & \textbf{85.8 (+7.5)}   & 50.3 (+3.6)            &\textbf{ 50.2 (+3.2)}      \\
		32                      & 91.8 (+8.0)          & 91.0 (+5.2)    & 48.5 (-1.8)            & 49.2 (-1.0) \\
		64                      & 95.0 (+4.8)          & 93.9 (+2.9)    & \textbf{53.1 (+3.6) }           & 51.8 (+2.6)     \\
		128                     & 96.4 (+1.4)           & 95.8 (+1.9)    & 54.3 (+3.2)            & 55.3 (+3.5)    \\
		160                     & 96.3 (-0.1)        & 95.7 (-0.1)    & 54.3 (0.0)           & 54.5 (-1.2)     \\ \hline
	\end{tabular}
}
\caption{The comparison of different K-shot Setting. We use the average of accuracy and the gain of accuracy between different k-shot setting to select the best k-shot setting. }
\label{Tab:B4}
\end{table*}

\section{Meta-prompt for Different tasks}
A compact Meta-prompt is consists of object description (OD), summary description (SD), and task description (TD). In the evaluation of Meta-prompt, all the Meta-prompt used in the proposed paper are list in Tab. \hyperref[Tab:C1]{C.1.}. Note that the label mappings in Meta-prompt still use the `Label Mapping' in Tab. \hyperref[Tab:B1]{B.1.}   
\label{sec:appendixC}
\setcounter{table}{0}
\renewcommand{\thetable}{C.\arabic{table}}
\begin{table*}[htb]
	\centering
	\begin{tabular}{l}
		\toprule
		\textbf{Task name (labels)}\\  -Metadata description template ([OD][SD][TD])\\                   
		\midrule
		\textbf{SST-2 } (positive / negative) \\ - A movie review \\ talking about its director, actor, performance, character skill, and story,\\ the emotion of this review was [MASK]  \\
		\midrule
		\textbf{SST-5} (very positive / positive / neutral / negative / very negative)  \\ - A movie review \\ talking about its director, actor, performance, character skill, and story,\\ the emotion of this review was [MASK]  \\
		\midrule
		\textbf{TREC}  (abbreviation / entity / description / human / location / numeric)  \\  	
		- A English question.  \\
		about huaman, description, location numeric entity , and abbreviations. \\
		The question type is [mask].  \\
		\midrule
		\textbf{SNLI } (entailment / neutral / contradiction)  \\ 
		- $[sent_{1}.]$  It is a Stanford Natural Language Inference sentence pairs \\	
		manually labeled as entailment , contradiction , and neutral. \\	
		whether the context contains the answer to the question?[mask]$[sent_2.]$ \\
		\midrule
		\textbf{QNLI} (entailment / contradiction)\\		
		- $[sent_{1}.]$  It is a Stanford Question Answering sentence pairs \\	
		manually labeled as entailment and contradiction. \\	
		whether the context contains the answer to the question? [mask]$[sent_2.]$
		\\
		\bottomrule
	\end{tabular}
	\caption{Our Meta-prompt combining by object description (OD), summary description (SD), and task description (TD) with a mask token.}
	\label{Tab:C1}
\end{table*}

\section{Multi-labels Mapping in ML-prompt}
\setcounter{table}{0}
\renewcommand{\thetable}{D.\arabic{table}}

\subsection{Single Multi-labels mapping}
\label{sec:appendixD}
Our proposed ML-prompt try to enlarge the mapping tokens in different label mapping to enhance the flexibility in the definition of label mapping, instead of mapping one most representative token for each label since it may need force searching to find this token in a huge vocabulary. The detail of Multi-labels mapping is shown in Tab. \hyperref[Tab:D1]{D.1.}. Note that the template in this section still use the original template in Tab. \hyperref[Tab:B1]{B.1}.
\begin{table*}[htb]
		\begin{tabular}{l p{15cm}}
			\hline
			\textbf{Task} & \textbf{Label Mapping}                                                                            \\ \hline
			SST-2         & \textbf{Negative}: terrible pathetic bad ridiculous , \textbf{Positive}: wonderful delicious gorgeous delightful.                                                               \\ \hline
			SST-5         &  \textbf{V. negative} :  disgusting awful horrible ridiculous boring , \textbf{Negative}:  simple better unnecessary weird doomed , \textbf{Neutral}:  hilarious hilarious dark weird dark , \textbf{Positive}: fascinating remarkable gorgeous incredible remarkable , \textbf{V. positive}:  magnificent wonderful terrific magnificent spectacular.                     \\ \hline
			SNLI         &  \textbf{Contradiction} : No inconformity inconsistent disagreement not , \textbf{Entailment} : Yes agree same accord good , \textbf{Neutral} : Maybe like seem might sound.             \\\hline
			QNLI          &  \textbf{Not\_entailment} : No Nonetheless Yet Notably , \textbf{Entailment} : Yes Okay Notably good.                                       \\ \hline
			TREC          &  \textbf{Description}: Description is depict what how , \textbf{Entity}: Entity concept Which name type , \textbf{Abbr.}: stand expression form does What , \textbf{Human}: Human team group organization man , \textbf{Loc}: Location street road city where , \textbf{Num}: Number How one many quantity.                     \\ \hline
		\end{tabular}
	\caption{The Multi-labels mapping in the ML-prompt.}
	\label{Tab:D1}
\end{table*}

\subsection{Different Multi-labels mappings}
By increasing the mapping tokens of each label to construct a series of multi-labels mappings, we explore the influence of different numbers of mapping tokens. We choose SST-5 sentiment classification task and QNLI text entailment task to evaluate the effectiveness of our proposed ML-prompt using different Multi-labels mappings. The detail label mappings are depicted in Tab. \hyperref[Tab:D2]{D.2.} and  Tab. \hyperref[Tab:D3]{D.3.}

\begin{table*}
	\centering
	\begin{tabular}{p{11cm}p{2cm}c}
		\toprule
		\textbf{Label mapping of QNLI}   & \textbf{QNLI}   & \textbf{Mapping Number}  \\
		\midrule
		\textbf{not\_entailment}: Fortunately And \\
		\textbf{entailment}: Recently Together    & 2   & 62.6 (6.7)   \\\midrule
		\textbf{not\_entailment}: Fortunately And Nonetheless \\ \textbf{entailment}: Recently Together Okay           & 3  & 66.3 (5.2)   \\\midrule
		\textbf{not\_entailment}: Fortunately And Nonetheless However \\ \textbf{entailment}: Recently Together Okay Yes    & 4  & 65.5 (5.1) \\\midrule
		\textbf{not\_entailment}: Fortunately And Nonetheless However Instead \\ \textbf{entailment}:  Recently Together Okay Yes So    &5  & 66.1 (5.3) 
		\\\midrule
		\textbf{Esemnble}  & - & \textbf{68.6 (3.7)} \\
		\bottomrule
		
	\end{tabular}
	\caption{The average accuracy and variance of different mapping with increasing tokens in label mapping of Ml-prompt. The result of `Ensemble' tag means that we also make a joint prediction with above-mentioned multi-labels mappings.}
	\label{Tab:D2}
\end{table*}

\begin{table*}
	\centering
	\begin{tabular}{p{12cm}ccccc}
		\toprule
		\textbf{Label mapping of SST-5}      & \textbf{SST-5 (manual)} \\
		\midrule
		\textbf{v.Neg.}: terrible disgusting , \textbf{Neg.}: bad simple , \textbf{netural}: okay neutral , \textbf{Pos.}: good better ,\textbf{ v.Pos.}: great fascinating        & 41.7(4.8)   \\\midrule
	\textbf{	v.Neg.}: terrible disgusting awful , \textbf{Neg.}: bad simple ordinary , \textbf{netural}: okay neutral fine, \textbf{Pos.}: good better pretty , \textbf{v.Pos.}: great fascinating magnificent             & 40.1 (4.6)  \\\midrule
		\textbf{v.Neg.}: terrible disgusting awful horrible , \textbf{Neg.}: bad simple ordinary ill , \textbf{netural}: okay neutral fine regular , \textbf{Pos.}: good better pretty gorgeous , \textbf{v.Pos.}: great fascinating magnificent wonderful        & 41.0(2.7)  \\\midrule
		\textbf{v.Neg}: terrible disgusting awful horrible ridiculous , \textbf{Neg.}: bad simple ordinary ill doomed , \textbf{netural}: okay neutral fine regular normal , \textbf{Pos.}: good better pretty gorgeous beautiful , \textbf{v.Pos.}: great fascinating magnificent wonderful spectacular                  & 40.7(1.7)
		\\\midrule
		\textbf{Esemnble}   & 43.5 (4.0)\\
		\bottomrule
		
	\end{tabular}
	\caption{The average accuracy and variance of different mapping with increasing tokens in label mapping of Ml-prompt. The result of `Ensemble' tag means that we also make a joint prediction with above-mentioned multi-labels mappings.}
	\label{Tab:D3}
\end{table*}

\end{document}